\documentclass[10pt,twocolumn,letterpaper]{article}

\usepackage{wacv}
\usepackage{times}
\usepackage{epsfig}
\usepackage{graphicx}
\usepackage{amsmath}
\usepackage{amssymb}
\usepackage{booktabs}
\usepackage{graphicx}
\usepackage{tikz}
\usepackage{comment}
\usepackage{amsmath,amssymb} % define this before the line numbering.
\usepackage{color, colortbl}

\usepackage{epsfig}
\usepackage{import}
\usepackage{booktabs}
\usepackage{multirow}
\usepackage{caption}
\usepackage{subcaption}
\usepackage{pifont}
\usepackage{multirow}
\usepackage{xspace}
\usepackage{epstopdf}
\usepackage{float}
\usepackage[export]{adjustbox}
\usepackage{dsfont}
\usepackage[nice]{nicefrac}
\usepackage{mathtools} % https://www.overleaf.com/learn/latex/Articles/Mathtools_-_for_beautiful_math

\usepackage{blindtext}

\usepackage[accsupp]{axessibility}

\makeatletter
\@namedef{ver@everyshi.sty}{}
\makeatother
\usepackage{tikz}

\newcommand\customparagraph[1]{\vspace{0.4em}\noindent\textbf{#1}}

\definecolor{_teal3}{HTML}{e6f5f0}
\newcommand{\putalg}{\textsc{LC}-loss\xspace}
\newcommand{\ours}{\textsc{Ours}\xspace}

\colorlet{seccolor}{black!70}
\newcommand\boldsec[1]{\textcolor{seccolor}{\textbf{#1}}}

\def\wacvPaperID{1035} % Enter the WACV Paper ID here

\wacvalgorithmstrack   % Uncomment this line if you are submitting to the Algorithms Track.
\wacvfinalcopy % *** Uncomment this line for the final submission

\ifwacvfinal
\usepackage[breaklinks=true,bookmarks=false]{hyperref}
\else
\usepackage[pagebackref=true,breaklinks=true,colorlinks,bookmarks=false]{hyperref}
\fi

\usepackage[capitalize]{cleveref}

\begin{document}

%%%%%%%%% TITLE

\title{Self-supervised Learning with Local Contrastive Loss for Detection and Semantic Segmentation}

\author{Ashraful Islam\thanks{This work was done while the author was an intern at Microsoft.}\\
Nvidia\\
{\tt\small aislam@nvidia.com}
\and
Ben Lundell, Harpreet Sawhney, Sudipta N.~ Sinha\\
Microsoft Mixed Reality\\
{\tt\small \{Benjamin.Lundell,Harpreet.Sawhney,Sudipta.Sinha\}@microsoft.com}
\and
Peter Morales \\
{\tt\small io.peter.morales@gmail.com}
\and 
Richard J.~Radke\\
Rensselaer Polytechnic Institute \\
{\tt\small rjradke@rpi.edu}
}

\maketitle
\thispagestyle{empty}

%%%%%%%%% ABSTRACT

\begin{abstract}
    We present a self-supervised learning (SSL) method suitable for semi-global tasks such as object detection and semantic segmentation. We enforce local consistency between self-learned features that represent corresponding image locations of transformed versions of the same image, by minimizing a pixel-level local contrastive (LC) loss during training. LC-loss can be added to existing self-supervised learning methods with minimal overhead. We evaluate our SSL approach on two downstream tasks -- object detection and semantic segmentation, using COCO, PASCAL VOC, and CityScapes datasets. Our method outperforms the existing state-of-the-art SSL approaches by 1.9\% on COCO object detection, 1.4\% on PASCAL VOC detection, and 0.6\% on CityScapes segmentation.
\end{abstract}

%%%%%%%%% BODY TEXT

% \vspace{-2mm}
\section{Introduction}
% \vspace{-2mm}

% no one cites the fukushima paper for SSL -- so I don't know why we are doing that?
% \cite{fukushima1980neocognitron}.

Self-supervised learning (SSL) approaches learn generic feature representations from data in the absence of any external supervision. These approaches often solve an \textit{instance discrimination} pretext task in which multiple transformations of the same image are required to generate similar learned features.
\begin{comment}
SSL approaches often solve an \textit{instance discrimination} pretext task in which multiple transformations of the same image are required to generate similar learned features~\cite{chen2020simplesimclr,he2020momentummoco}.
SSL remains an active area in vision  research~\cite{caron2020unsupervisedswav,caron2021emergingdino,grill2020bootstrapbyol,he2020momentummoco}.
\end{comment}
Recent SSL methods have shown remarkable promise in {\em global} tasks such as classifying images by training simple classifiers on the features learned via instance discrimination~
~\cite{caron2020unsupervisedswav,caron2021emergingdino,chen2020simplesimclr,grill2020bootstrapbyol,he2020momentummoco}. However, global feature-learning SSL approaches do not explicitly retain spatial information thus rendering them ill-suited for semi-global tasks such as object detection, and instance and semantic segmentation \cite{wang2021densecl,xie2021propagatepixpro}.
\begin{comment}
Recent SSL methods have shown remarkable promise in global tasks such as classifying images by applying simple linear classifiers on the features learned via instance discrimination~
\cite{chen2020simplesimclr,grill2020bootstrapbyol,he2020momentummoco}.
BYOL~\cite{grill2020bootstrapbyol} and SwaV~\cite{caron2020unsupervisedswav} are two such examples that train ResNet-50 features~\cite{he2016deepresnet} that resulted in 74.3\% and 75.3\% top-1 accuracy on ImageNet whereas the top-1 accuracy achieved by training ResNet-50 features in a fully supervised setting was 76.6\%.
\end{comment}

\begin{figure}[!t]
    \centering
    \includegraphics[width=0.8\linewidth]{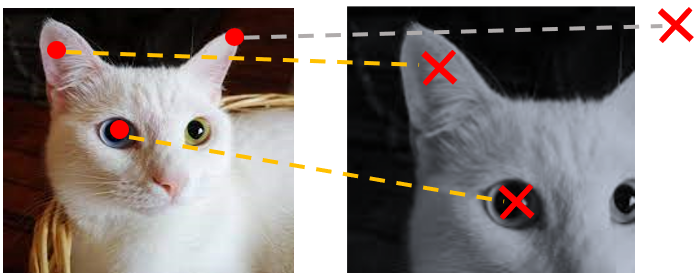}
    \vspace{-2.4mm}
    \caption{\small Our framework encourages local regions of two transformed images to learn similar features. The first image (left) uses color augmentation only, and another image (right) employs both spatial (random resize crop) and color transformations. With known corresponding pixels a consistency loss enforces maximal similarity between the corresponding learned features.
        % \sudipta{Figure and caption needs work -- we know other methods also impose local consistency constraints. Furthermore we say 'we propose a new loss that' but the figure is not neither explaining what the new loss function is or why it is new. The red circles, crosses and white lines must be explained.}
    }
    % \vspace{-4mm}
    % \sudipta{make figures side-by-side.}
    %The corresponding pairs are automatically obtained via data augmentation.
    \label{fig:teaser}
    \vspace{-4mm}
\end{figure}

\begin{comment}
The loss functions used by existing SSL methods to train convolutional neural network backbones are well suited for learning global features that are effective for the image categorization task. However, such representations do not retain spatial information. This makes them less suitable for semi-global tasks such as object detection, instance segmentation and semantic segmentation.
\end{comment}
This work focuses on extending SSL to incorporate spatial locality by using a local contrastive (LC) loss function at a dense and fine-grained pixel level. The main idea is illustrated in Figure \ref{fig:teaser}. Specifically, we encourage corresponding {\em local pixels} in the two transformed images to produce similar features. The true pixel correspondences are known since the input image pairs are generated by applying two distinct transformations to a single image. Note that this approach can be used along side {\em any} conventional global SSL objective with minimal overhead.

We evaluate the impact of LC-loss on several downstream tasks, namely object detection, and instance and semantic segmentation and report promising improvements over previous spatially-aware SSL methods \cite{wang2021densecl,wei2021aligningsoco,xiao2021regionresim,xie2021propagatepixpro} on Pascal VOC, COCO and Cityscapes datasets.

\customparagraph{Contributions.} Our main contribution is in demonstrating that adding a pixel-level contrastive loss to the BYOL~\cite{grill2020bootstrapbyol} training procedure for the instance discrimination pretext task is sufficient to produce excellent results on many downstream dense prediction tasks. A similar, pixel-level contrastive loss formulation was presented in PixPro \cite{xie2021propagatepixpro}, but a more complicated pixel-to-propagation consistency pre-text task was required to achieve state of the art results in dense prediction tasks. We show that no additional pre-text is necessary, and our simpler local contrastive loss formulation achieves superior performance. Specifically, our key technical contributions are: \textbf{(1)} a simple framework that computes local contrastive loss (\putalg) to make the corresponding pixels of two augmented versions of the same image similar, that can be added to any self-supervised learning method, such as BYOL; and \textbf{(2)} state-of-the-art transfer learning results in several dense labeling tasks. Using ResNet-50 backbones pretrained on ImageNet, our BYOL variant achieves 40.6 AP for COCO object detection (+1.9 vs SSL SOTA\footnote{State-Of-The-Art}), 60.1 AP for VOC object detection (+1.4 vs SOTA), 72.1 mIoU for VOC segmentation (+1 vs SOTA), and 77.8 mIoU for CityScapes segmentation (+0.6 vs SOTA) for full-network fine-tuning setting.  Our performance improvement is even more significant in the frozen backbone setting discussed in Section~\ref{sec:analysis}.

\section{Related Work} \label{sec:literature}
\vspace{-2mm}
\customparagraph{Self-supervised learning (SSL).}
In self-supervisd learning, the supervisory signal is automatically generated from a pair of input images and a pretext task. The input pair is generated by applying two distinct transformations. The pretext task is comparison between the learned representations of each pair of input images . Various pretext tasks have been explored, such as, patch position \cite{Doersch_2015_ICCV}, image colorization \cite{zhang2016colorful}, image inpainting \cite{pathak2016context}, rotation \cite{gidaris2018unsupervised}, and predictive coding \cite{henaff2019datacpc}. The pretext task that has shown the most promise is the instance discrimination task, in which each image is considered as a single class. SimCLR \cite{chen2020simplesimclr}, the first to propose this pretext task, adopts contrastive learning in which features from augmented versions of the same image are made closer in the feature space than all the other images in a mini-batch. SimCLR requires a large mini-batch to make contrastive learning feasible. MoCo \cite{he2020momentummoco} solves the issue of large batch size using a momentum queue and moving average encoder. Despite impressive results in image classification tasks, contrastive learning requires careful handling of negative pairs.  Recent approaches like BYOL \cite{grill2020bootstrapbyol}, SwaV \cite{caron2020unsupervisedswav}, and DINO \cite{caron2021emergingdino}  do not require any negative pairs or a memory bank. They also achieve impressive performance on the ImageNet1k linear evaluation task and downstream image classification-based transfer learning tasks.

\customparagraph{SSL for Detection and Segmentation.}
For dense prediction tasks, SSL methods use an ImageNet-pretrained backbone within a larger architecture designed for a detection or segmentation task \cite{long2015fullyfcn,ren2016fasterrcnn,xiao2018unifiedupernet}, and fine tune the network on the downstream task dataset. He et al.~\cite{he2019rethinking}reported that ImageNet pretrained models might be less helpful if the target task is localization sensitive than classification . One potential solution is to increase the target dataset size \cite{he2019rethinking}, or to impose local consistency during the ImageNet pretraining. We adopt the latter strategy.

Broadly, there are two approaches in the literature to ensuring local consistency during self-supervised pretraining: pixel-based and region-based. In region-based methods, first region proposals are generated - either during the self-supervised training \cite{wei2021aligningsoco,roh2021spatiallyscrl,yang2021instanceinsloc}, or before training starts \cite{xiao2021regionresim}, and then local consistency is applied between pooled features of the proposed regions. Our approach is pixel-based; local consistency is applied between the local features for corresponding pixels of transformed versions of the same image \cite{wang2021densecl,xie2021propagatepixpro,xie2021detco}.

% \begin{figure*}[!htbp]
%     \vspace{-2mm}
%     \centering
%     \begin{subfigure}[t]{0.28\textwidth}
%         \centering
%         \includegraphics[width=\textwidth]{Figures/densecl.png}
%         \caption{DenseCL \cite{wang2021densecl}}
%         \label{diff_dense}
%     \end{subfigure}
%     \begin{subfigure}[t]{0.3\textwidth}
%         \centering
%         \includegraphics[width=\textwidth]{Figures/pixpro.png}
%         \caption{PixPro \cite{xie2021propagatepixpro}}
%         \label{diff_pix}
%     \end{subfigure}
%     \begin{subfigure}[t]{0.3\textwidth}
%         \centering
%         \includegraphics[width=\textwidth]{Figures/ours.png}
%         \caption{\ours}
%         \label{diff_ours}
%     \end{subfigure}
%     \vspace{-2mm}
%     \caption{{\bf Contrasting DenseCL, PixPro, and \ours}. {\small C: color transformations (blurring, color jitter, etc.), C+S: both spatial (random crop) and color transformations. (a) DenseCL calculates corresponding pixels by maximizing similarity of local features between transformed images. (b) Both PixPro and \ours pre-calculate matching pixels. PixPro warps the features of the transformed images to the original image, and applies a pixel-propagation module to measure local contrastive loss. (c) we simply calculate the log-similarity of corresponding pixels by applying bi-linear interpolation after measuring the correspondence.}}
%     \label{fig:diff}
% \end{figure*}

DenseCL \cite{wang2021densecl} proposed dense contrastive learning for self-supervised visual pretraining. It follows the MoCo \cite{he2020momentummoco} framework to formulate the dense loss. However, DenseCL does not use known pixel-correspondences to generate positive pairs of local feature between two images. Instead, it extracts the correspondence across views. This creates a chicken-and-egg problem where DenseCL first requires learning a good feature representation to generate correct correspondences.

Our local contrastive loss is more similar to the PixContrast loss in the PixPro paper \cite{xie2021propagatepixpro}. Given an image $I$, both methods operate on two distinct transforms $J_1 = \mathcal{T}_1(I)$ and $J_2 = \mathcal{T}_2(I)$ of $I$ to produce two low-resolution, spatial feature maps. Both methods use a contrastive loss using pixel correspondences to generate positive and negative samples. The difference comes in how these samples are selected.

In PixContrast, pixels in the low-resolution feature map are warped back to the original image space using $\mathcal{T}_1^{-1}$ and $\mathcal{T}_1^{-1}$. Positive samples are then determined by all pairs of pixels that are sufficiently close after warping. Our method generates positive samples using the correspondences in $J_1$ and $J_2$ derived directly from $\mathcal{T}_1$ and $\mathcal{T}_2$. While similar to ours, the method proposed for PixContrast does not work. Instead, an additional pixel-propagation module is introduced (PixPro) to measure the feature similarity between corresponding pixels. We show that no pixel propagation or feature warping is required in our simpler formulation. %(see Figure \ref{diff_ours}).

In summary, our framework does not require: (1) an encoder-decoder architecture for local correspondence loss \cite{pinheiro2020unsupervisedVADeR}; (2) contrastive learning that needs carefully tuned negative pairs \cite{pinheiro2020unsupervisedVADeR,wang2021densecl,xiao2021regionresim}; (3) a good local feature extractor to find local feature correspondences \cite{wang2021densecl}; and (4) an additional propagation module to measure local contrastive loss \cite{xie2021propagatepixpro}. A simple local correspondence loss obtained from matching pixel pairs achieves state-of-the-art results in detection and segmentation tasks.

\section{Methodology}
% \vspace{-2mm}
% There are many approaches for self-supervised representation learning in the literature. In this paper, we adopt BYOL \cite{grill2020bootstrapbyol} as the default method. BYOL achieves higher performance than contrastive learning without using any negative pairs, and is more resilient to changes in hyper-parameters like batch size and augmentation. 

\begin{figure*}[!tbp]
    \centering
    \includegraphics[width=0.8\textwidth]{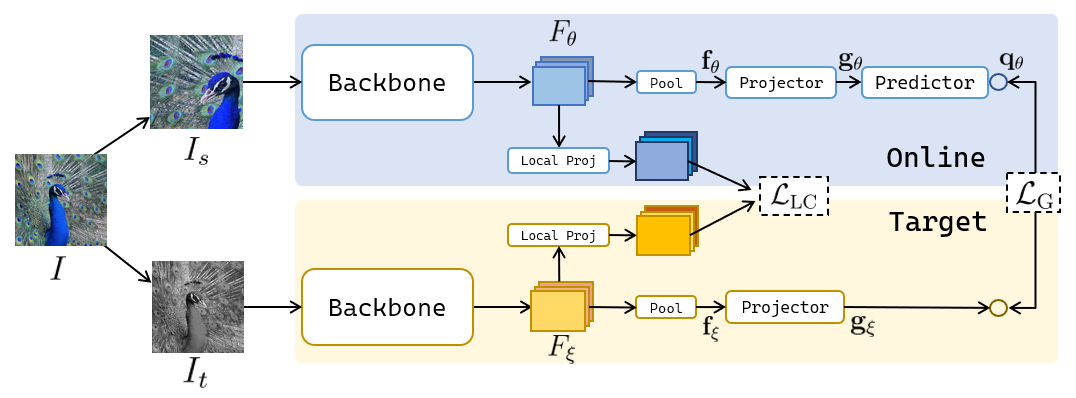}
    \vspace{-2mm}
    \caption{\small {\bf Proposed Framework.} We use BYOL  as the  self-supervised learning framework. BYOL consists of an online network with parameters $\theta$ and a target network with parameters $\xi$ which are the exponential moving average of $\theta$. Given an image, we create two transformed versions. We apply mean squared error loss between the L2-normalized global feature representations from the online and target networks.. We also calculate a local contrastive loss from the dense feature representations of the image pairs.
        % Fig \ref{fig:lcl} illustrates how local loss is obtained.
        % \sudipta{caption is not explaining very well what is shown in the figure. What is $\mathcal{L}_{SS}$?}
    }
    \vspace{-2mm}
    \label{fig:pipeline}
\end{figure*}

Figure \ref{fig:pipeline} depicts our LC-loss framework. We use BYOL for the global self-supervised loss function, and apply a local contrastive loss on dense feature representations obtained from the backbone networks. We adopted BYOL framework because it achieves higher performance than contrastive learning without using any negative pairs, and it is more resilient to changes in hyper-parameters like batch size and image transformations. In the following, we briefly describe the BYOL framework, and introduce our approach.

\customparagraph{Instance Discrimination from Global Features.}
BYOL consists of two neural networks: an online network with parameters $\theta$ and a target network with parameters $\xi$. The online network has three sub-networks: an encoder, a projector and a predictor. The target network has the same architecture as the online network except the predictor. The online network is updated by gradient descent, and the parameters of the target network are exponential moving averages of the parameters of the online network. Given an input image $I$, two transformed views $I_c$ and $I_t$ of $I$ are obtained.
% \sudipta{Why are you using so many different sets of notations -- $v$, $v'$ and $x$ here and $I_c$, $I_t$ later and then afterwards $I_c$, $I_{sc}$, and $T_{sc}$? Just use consistent notation everywhere. Call the images $I$ and $J$ and the transformation $T$.} 
One  view $I_c$ is passed through the online network with parameters $\theta$ to obtain local features $F_{\theta}$, average pooled encoder output ${\bf f}_{\theta}$, a projection ${\bf g}_{\theta}$ and prediction ${\bf q}_{\theta}$. View $I_t$ is passed through the target network with parameters $\xi$ to obtain local features $F_{\xi}$, average pooled encoder output $f_{\xi}$ and a projection ${\bf g}_{\xi}$. Check Fig.~\ref{fig:pipeline}. There is no predictor in the target network. This asymmetric design is adopted to prevent collapse during self-supervised training \cite{grill2020bootstrapbyol}. We follow the original BYOL network to set the dimension of encoder outputs (2048 dim), projections (4096 dim), and predictions (256 dim). 
% \sudipta{Why are we taking dotproducts of a vector q and another one which is denoted g? Why are we not denoting both vectors as q -- of course one of them has a subscript theta and the other one will have a subscript xi.}
The global self-supervised loss function for a single input is defined as:

% \begin{equation}
%     \mathcal{L}_{\text{SS}} = 2 - 2 \ \text{cos}({\bf q}_{\theta}, {\bf q}_{\xi})
% \end{equation}

\begin{equation}
    \mathcal{L}_{\text{G}} = 2 - 2 \frac{{\bf q}_{\theta}^T {\bf g}_{\xi}}{
    ||{\bf q}_{\theta}||_2 ||{\bf g}_{\xi}||_2
    }
\end{equation}
% \sudipta{Let us make the vectors bold}
where ${\bf q}_{\theta}$ and ${\bf g}_{\xi}$ are learned global representations for two transformed images that are forced to be similar under cosine similarity.

\customparagraph{Local Contrastive Loss.}
Given two transformed versions of the same image,  namely $I_c$ and $I_t$, let an image point $p_c$ in $I_c$, correspond to another image point $p_t$ in $I_t$. We can determine $p_t$ for every $p_c$ given the known image transformations. The correspondence map $C_{p_c} \in \mathbb{R}^{H\times W}$ for source point ${p_c}$ is calculated, where $C_{p_c}(p_l)$ denotes the similarity score between $p_c$ of $I_c$ and every pixel $p_l$ of $I_t$. We define $C_{p_c}(p_l)$ to be the similarity score that $p_l$ is the corresponding pixel of $p_c$. As we know $p_t$ is the actual corresponding pixel, we want $C_{p_c}(p_t)$ to be maximized. The local loss for $p_c$ is the negative log likelihood at $p_t$ which encourages maximizing the likelihood estimate for the target locations $ -\log{C_{p_c}(p_t)}$.
% }

% \sudipta{The text above marked in red is again repeatedly later in the next page with all the equations. Bring all that math, equations etc. here and describe the math and the idea, everything to do with LC-loss over here (make it a subsection, Section 3.1). Finally, after Section 3.1, end with a Section 3.2 but I am not even sure why we should call it a "pipeline". Anyway, we will figure that later... but all the details about how we generate transformations, correspondences, bilinear interpolation, let's move all that to the end.}

% \customparagraph{Proposed Pipeline.}
% \todo{Ashraful update Figure 2 to remove implementation details, annotate with parameters, and include reference to it here.}
We now describe how this is incorporated in our pipeline. We employ learned feature-level correspondence as a measure of pixel-level correspondence. Given an image $I$, we apply image transformation $\mathcal{T}_{\text{c}}$ to $I$ to obtain $I_{\text{c}}$. $\mathcal{T}_{\text{c}}$ contains strong color transformations (for example, Gaussian blur, solarization, color distortion), but does not include spatial transformations like image flipping or random crop. We apply normal resize operation to resize the image to $H\times W$ shape. Another transformation $\mathcal{T}_{\text{sc}}$ is applied to $I$ to obtain $I_{\text{sc}}$, where $\mathcal{T}_{\text{sc}}$ contains both spatial and color transformations. We obtain dense feature representations of $I_{\text{sc}}$ from the backbone of the online network, and denote it as $F_{\theta} \in \mathbb{R}^{h\times w}$ where $h=\nicefrac{H}{p}$ and $w=\nicefrac{W}{p}$ with $p$ as the stride size for the feature representation. We get a similar feature representation $F_{\xi} \in \mathbb{R}^{h\times w}$ by passing $I_{\text{c}}$ through the target network.

Next we select $h \times w$ image points on a 2D uniform grid in $I_{\text{c}}$. Each point $p_{\text{c}}$ of $I_{\text{c}}$ has the local feature representation $F_{\xi}(p_{\text{c}})$ obtained from the target network. For the point $p_{\text{c}}$, there is a corresponding point $p_{\text{sc}}$ in $I_{\text{sc}}$. Note that because of random crop and flipping in $\mathcal{T}_\text{sc}$,  the feature $F_{\theta}(p_{\text{sc}})$ may not be at an integer pixel coordinate $p_{\text{sc}}$. Instead of adopting an expensive high-dimensional warping of $F_{\theta}$ to obtain corresponding features, we compute the negative log-likelihood from correspondence map and resample the negative log-likelihood using bilinear interpolation to deal 2D points with subpixel coordinates, which is described below \cite{germain2021visualnre}.

We compute a dense correspondence map $C' \in \mathbb{R}^{(h \times w) \times (h \times w)}$ between $I_{\text{c}}$ and $I_{\text{sc}}$ as,
% \begin{equation}
%     C'(p_c, p_t) = \text{cos}(F_{\xi}(p_c), F_{\theta}(p_t))
% \end{equation}
\begin{equation}
    C'(p_c, p_t) = \frac{F_{\xi}(p_c)^T F_{\theta}(p_{\text{sc}})}{||F_{\xi}(p_c)||_2 ||F_{\theta}(p_{\text{sc}})||_2}
\end{equation}
where $p_c$ and $p_t$ are image points from  $I_{\text{c}}$  and  $I_{\text{sc}}$ respectively.
% Assuming $p_{\text{c}}$ is a keypoint in $I_{\text{c}}$, and $p_t$ is any keypoint in $I_{\text{sc}}$, We measure the correspondence map between $F_{\xi}(p_{\text{c}})$ and $F_{\theta}(p_{t})$, denoted as $C' \in \mathbb{R}^{(h \times w) \times (h \times w)}$, by 
% \begin{equation}
%     C'(p_c, p_t) = \text{cos}(F_{\theta}(p_c), F_{\xi}(p_t))
% \end{equation}
% Next, we divide each element of $C'$ by local temperature $\tau$ and apply a softmax operation along the last dimension to get a likelihood estimation $C \in \mathbb{R}^{(h \times w) \times (h \times w)}$. 
Then, we calculate the negative log-likelihood

\begin{equation}
    \text{NLL}(p_{\text{c}}, p_\text{t}) = -\log\frac{
        \exp(C'(p_c, p_{\text{sc}}) / \tau)
    }{
    \sum_{\substack{k \in \Omega_{sc}}} (\exp(C'(p_c, p_k) / \tau) )
    }
\end{equation}
where $\Omega$ are set of locations in the $I_{sc}$.
% $\text{NLL}(p_c, p_t) = -\log(C(p_c, p_t))$. 
If $p_{sc}$ is not an integer location, we obtain the negative log-likelihood $\text{NLL}(p_{\text{c}}, p_\text{sc})$ by bilinearly interpolating $\text{NLL}(p_{\text{c}}, :)$. The contrastive loss, \putalg, is defined as
\begin{equation}
    \mathcal{L}_{\text{LC}} = \frac{1}{|\mathcal{P}|} \sum_{\substack{
            (p_{\text{c}}, p_{\text{sc}}) \in \mathcal{P}
        }}
    \text{NLL}(p_{\text{c}}, p_{\text{sc}})
\end{equation}
where $\mathcal{P}$ contains all corresponding pairs $\{(p_{\text{c}}, p_{\text{sc}})\}$ for $I_{\text{c}}$ and $I_{\text{sc}}$ such that $p_{\text{sc}}$ does not fall outside of the boundary of $I_{\text{sc}}$, and $|\mathcal{P}| \le h \times w$.

% \sudipta{We don't need a separate equation for NLL(*,*) = .... Just add the -log(C(*,*)) term directly into the next equation. However, we need an equation to define C(*,*), i.e. explain how you get C(*,*) from C'(*,*) mathematically.}

% \sudipta{In the formula for NLL, in the text above, there is no soft-max. Do we use softmax like everyone else, or do we not? If the answer is yes we do, then the equations are incorrect. Softmax is the function that applies exponentiation to the scores in $C(\cdot, \cdot)$ and then normalizes them, so that the values sum to 1.The other papers DenseCL, PixPro, both show that equation but we don't. In what we show we are missing the exponentiation and the normalization.} \ash{I mentioned the softmax part in the text, and the C(,) contains normalized score after performing the softmax operation. I think it looks confusing, and I will write it down in equation form.}

Our total loss is defined as:

\begin{equation} \label{eqn:loss_total}
    \mathcal{L} = (1-\alpha)\mathcal{L}_{\text{G}} + \alpha \mathcal{L}_{\text{LC}}
\end{equation}
where $\alpha$ is the multiplicative factor that balances the two loss components. See Section~\ref{sec:ablation} for a study on the impact of $\alpha$.
% Optimizing for $\mathcal{L}_{\text{G}}$ results in better global features for image classification tasks, whereas optimizing $\mathcal{L}_{\text{LC}}$ learns better spatial representations for dense tasks like detection or segmentation.\ben{That's not actually the conclusion we reach during the ablation study. So we should just say, "See Section (whatever) for a study on the impact of the parameter $\alpha$.}
% \sudipta{Can we shorten the subscripts, $\mathcal{L}_{G}$ instead of $\mathcal{L}_{\text{G}}$ and $\mathcal{L}_{\text{LC}}$ instead of $\mathcal{L}_{\text{LC}}$?}
\section{Experiments}
% \vspace{-2mm}
\subsection{Implementation Details}
\customparagraph{Pretraining Setup.}
We use ResNet-50 \cite{he2016deepresnet} as the backbone network and BYOL \cite{grill2020bootstrapbyol} as the self-supervision architecture. We use identical architectures for projection and prediction networks as in BYOL. For extracting local features, we add a local projection branch on the online network and another branch of similar architecture on the target network. As with the global branches, only the online local projection branch is updated through optimization while the target local projection branch is the exponential moving average of the online one. The local projection branch consists of two convolution layers. The first convolution layer consists of a $1\times 1$ convolution kernel with input dimension 2048 and target dimension 2048 following by a BatchNorm layer. The second convolution layer contains a $1\times 1$ kernel with output dimension 256. The input to the local projection branch is the local feature representation from the final stage of ResNet-50 (before the global average pooling layer). % \customparagraph{Transformations.}
For image transformations during pretraining, following BYOL \cite{grill2020bootstrapbyol}, we use random resize crop (resize to $224\times 224$), random horizontal flip, color distortion, blurring, and solarization. We do not apply random crop for the image that is used to obtain local contrastive loss.

\customparagraph{Dataset.}
We use the ImageNet \cite{deng2009imagenet} dataset for pretraining the networks. ImageNet contains $\sim$1.28M training images, mostly with a single foreground object.

\customparagraph{Optimization.}
The default model is trained with 400 epochs if not specified in the results. See Sec.~\ref{sec:ablation} for details on the effect of pretraining epoch to the transfer performance. The LARS optimizer is used with a base learning rate of 0.3 for batch-size 256, momentum 0.9, weight-decay 1e-6, and with cosine learning rate decay schedule for with learning rate warm-up for 10 epochs. We use 16 GPUs with 256 batch-size on each GPU, hence, the effective batch-size is 4096. We linearly scale the learning rate with the effective batch size. The weight parameter $\alpha$ is set to 0.1 (Eqn.~\ref{eqn:loss_total}). For the momentum encoder, the momentum value starts from 0.996 and ends at 1. We use 16 bit mixed-precision during pre-training.

\subsection{Results on Object Detection and Instance Segmentation}
% \sudipta{Reorganize all the text in this subsection into three paragraphs.}
We use Detectron2 framework \cite{wu2019Detectron2} for evaluation of downstream object detection and segmentation results on COCO and PASCAL VOC dataset.

\begin{table*}[!htbp]
    \vspace{-3mm}
    \centering
    % \begin{minipage}[t]{0.7\textwidth}
    %\begin{minipage}[t]{1.0\textwidth}
    \begin{adjustbox}{max width=\linewidth}
        \begin{tabular}{l c | c c c | c c c | c c c}
            \toprule
            \multirow{4}{*}{\bf \small Method}      & {}                    & \multicolumn{9}{c}{\small \bf COCO}                                                                                                                                    \\
                                                    & {}                    & \multicolumn{3}{c|}{\small Object Detection} &
                                                    \multicolumn{3}{c|}{\small Object Detection} & \multicolumn{3}{c}{\small Instance Segmentation}                                                                        \\
                                                    & {\bf \small Pretrain} & \multicolumn{3}{c|}{\small RetinaNet + FPN}  & \multicolumn{3}{c|}{\small Mask-RCNN + FPN}  & \multicolumn{3}{c}{\small Mask-RCNN + FPN}                                                                              \\
                                                    & {\bf \small Epochs}   & { $\text{AP}^b$}                             & { $\text{AP}^b_{50}$}                            & { $\text{AP}^b_{75}$} & { $\text{AP}^b$}                             & { $\text{AP}^b_{50}$}                            & { $\text{AP}^b_{75}$} & AP$^{mk}$  & AP$^{mk}_{50}$ & AP$^{mk}_{75}$ \\
            \midrule

            Supervised  \cite{he2016deepresnet}     & 90                    & 37.7                                         & 57.2                                             & 40.4                  & 38.9           & 59.6           & 42.7      & 35.4       & 56.5           & 38.1
            \\
            Moco v2   \cite{chen2020improvedmocov2} & 200                   & 37.3                                         & 56.2                                             & 40.4      & 40.4           & 60.2           & 44.2                & 36.4       & 57.2           & 38.9           \\
            % BYOL  \cite{grill2020bootstrapbyol}     & 300                   & 35.4                                         & 54.7                                             & 37.4      & 37.2           & 58.8           & 39.8                   & 37.2       & 58.8           & 39.8
            BYOL  \cite{grill2020bootstrapbyol}     & 300                   & 35.4                                         & 54.7                                             & 37.4      & 40.4          & 61.6           & 44.1                   & 37.2       & 58.8           & 39.8
            \\
            \midrule
            DetCo        \cite{xie2021detco}        & 800                   & 38.4                                         & 57.8                                             & 41.2      & 40.1           & 61.0           & 43.9               & 36.4       & 58.0           & 38.9           \\
            ReSim-FPN \cite{xiao2021regionresim}    & 200                   & 38.6                                         & 57.6                                             & 41.6    & 39.8           & 60.2           & 43.5                  & 36.0       & 57.1           & 38.6           \\
            SCRL \cite{roh2021spatiallyscrl}        & 800                   & 39.0                                         & 58.7                                             & 41.9        & - & - & -           & 37.7       & 59.6           & 40.7           \\
            SoCo    \cite{wei2021aligningsoco}      & 400                   & 38.3                                         & 57.2                                             & 41.2       & \bf43.0        & {63.3} & 47.1           & 38.2       & { 60.2}     & {41.0}         \\
            \midrule
            DenseCL       \cite{wang2021densecl}    & 200                   & 37.6                                         & 56.6                                             & 40.2     & 40.3           & 59.9           & 44.3                 & 36.4       & 57.0           & 39.2           \\
            PixPro   \cite{xie2021propagatepixpro}  & 400                   & 38.7                                         & 57.5                                             & 41.6         & 41.4           & 61.6           & 45.4            & -          & -              & -              \\
            \rowcolor{_teal3} \ours                 & 400                   & \bf40.6                                      & 60.4                                          & 43.6    & 42.5           & 62.9           & 46.7          & {\bf 38.3} & {60.0}         & { 41.1}     \\
            \bottomrule
        \end{tabular}
    \end{adjustbox}
    % \end{minipage}
    % \hfill
    % \begin{minipage}[c]{0.28\textwidth}
    %\begin{minipage}[c]{1.0\textwidth}
    \vspace{-2mm}
    \caption{\small \textbf{Main Results.} We use faster-RCNN with RetinaNet for COCO object detection,  Mask-RCNN with FPN for COCO instance segmentation, Faster RCNN with FPN for VOC object detection. 
    % We use the reported scores from the literature for the existing methods.
    }
    \label{tab:main_ft}
    % \end{minipage}
    \vspace{-3mm}
\end{table*}

\customparagraph{COCO Object Detection.}
For object detection on COCO, we adopt RetinaNet \cite{lin2017focalretina}  following \cite{wei2021aligningsoco,roh2021spatiallyscrl,xie2021detco}. We finetune all layers with Sync BatchNorm for 90k iterations on COCO \texttt{train2017} set and evaluate on COCO \texttt{val2017} set. Table \ref{tab:main_ft} shows object detection results on COCO for our method and other approaches in the literature with full-network finetuning. Note that ReSim, SoCo, and SCRL use region proposal networks during pretraining on ImageNet, hence, these approaches are not exactly comparable to ours. Our model is more similar with methods like DetCo, DenseCL, and PixPro. We achieve 40.6 AP for object detection tasks outperforming the second best method PixPro \cite{xie2021propagatepixpro} by a significant 1.9\%. We also report results on COCO detection using Mask R-CNN + FPN. We again outperform PixPro (our mAP is higher by 1.4), when using Mask R-CNN + FPN for the detector.

\customparagraph{COCO Instance Segmentation.}
We use the Mask-RCNN framework \cite{he2017maskrcnn} with ResNet50-FPN backbone. We follow the $1\times$ schedule. Table \ref{tab:main_ft} depicts that we achieve 38.3\% AP for COCO instance segmentation, which is comparable with the SoCo \cite{wei2021aligningsoco}. Note that SoCo performs selective search on the input image to find object proposals, and uses additional feature pyramid networks during pre-training.

% Talk about results specific to COCO object detection. refer to Table~\ref{tab:coco_mrcnn} and discuss both frozen backbone and finetued backbone.

\customparagraph{PASCAL VOC Object Detection \cite{everingham2010pascalvoc}.}
We use the Faster-RCNN \cite{ren2016fasterrcnn} object detector with ResNet50-FPN backbone following \cite{roh2021spatiallyscrl}. For training, we use images from both \texttt{trainval07+12} sets and we evaluate only on the VOC07 \texttt{test07} set. We use the pre-trained checkpoints released by the authors for the backbone network, and fine tune the full networks on the VOC dataset. Table \ref{tab:main_results} shows that we achieve 60.1 AP for VOC detection. Our method improves mean AP by a significant 3.2\% over baseline BYOL, and outperforms the current SOTA PixPro by 1.4\% AP. The improvement is even more significant in AP75, where we outperform BYOL by 3.6\% and PixPro by 1.9\%.

\begin{table}[!htbp]
    \vspace{-2mm}
    \centering
    % \begin{minipage}[c]{0.55\textwidth}
    \begin{adjustbox}{max width=\linewidth}
        \begin{tabular}{l c |  c c c }
            \toprule
            \multirow{4}{*}{\bf \small Method}                 & {}                    & \multicolumn{3}{c}{\bf \small PASCAL VOC}                                                \\
                                                               & {}                    & \multicolumn{3}{c}{Object Detection}                                                     \\
                                                               & {\bf \small Pretrain} & \multicolumn{3}{c}{{ \small FRCNN + FPN}}                                                \\
                                                               & {\bf \small Epochs}   & { $\text{AP}^b$}                          & { $\text{AP}^b_{50}$} & {$\text{AP}^b_{75}$} \\
            \midrule

            Supervised  \cite{he2016deepresnet}                & 90                    & 53.2                                      & 81.7                  & 58.2                 %& 55.4              & 82.1                   & 60.9                      \\
            \\
            BYOL  \cite{grill2020bootstrapbyol}                & 300                   & 55.0                                      & 83.1                  & 61.1
            %& 56.9              & 83.4                   & 64.2               \\
            \\
            \midrule
            SCRL \cite{roh2021spatiallyscrl}                   & 800                   & 57.2                                      & 83.8                  & 63.9                 \\
            DenseCL$^{\dagger}$ \cite{wang2021densecl}         & 200                   & 56.6                                      & 81.8                  & 62.9                 \\
            PixPro$^{\dagger}$   \cite{xie2021propagatepixpro} & 400                   & 58.7                                      & 82.9                  & 65.9                 \\
            \rowcolor{_teal3} \ours                            & 400                   & \bf60.1                                   & \bf84.2               & \bf67.8              \\
            \bottomrule
        \end{tabular}
    \end{adjustbox}
    % \end{minipage}
    \hfill
    % \begin{minipage}[l]{0.4\textwidth}
    \vspace{-2mm}
    \caption{\small \textbf{Main Results.} We use Faster RCNN with FPN for VOC object detection. Supervised and BYOL results are from \cite{roh2021spatiallyscrl}. ($^{\dagger}$): We use pre-trained checkpoint released by the authors and fine tune on the VOC dataset.}
    \label{tab:main_results}
    % \end{minipage}
    \vspace{-3mm}
\end{table}

\subsection{Results on Semantic Segmentation}
We show semantic segmentation evaluation in Table \ref{tab:main_segm} on PASCAL VOC and  CityScapes \cite{cordts2016cityscapes} datasets for both fine tuning and frozen backbone settings. We use FCN backbone \cite{long2015fullyfcn} following the settings in \texttt{mmsegmentation} \cite{mmseg2020}.

\customparagraph{PASCAL VOC Segmentation.}
% \sudipta{Talk about relevant results. refer to Table~\ref{tab:coco_mrcnn} and discuss both frozen backbone and finetued backbone.}
We train on  VOC \texttt{train-aug2012} set for 20k
iterations and evaluate on val2012 set. Table \ref{tab:main_segm} shows that on the VOC2007 test set, our method yields 72.1\% mIoU outperforming BYOL by a 7.7\% and PixPro by 1\% mIoU.

\customparagraph{Cityscapes Segmentation.}
CityScapes \cite{cordts2016cityscapes} contains images from urban street scenes. Table \ref{tab:main_segm} shows that for fine tuning setting our approach yield 77.8\% AP which is 6.2\% mIoU improvement over BYOL and 0.6\% improvement over PixPro.

% \customparagraph{ScanNet Segmentation.}
% ScanNet is a dataset of RGB-D scans of real world environments from 1513 scans acquired in 707 distinct spaces; hence the dataset contains images from indoor scenes. We sample $\sim$20k images from 1201 scans as the training set and $\sim$5.5k images from 312 scans as the test set for semantic segmentation. On ScanNet, we achieve 55.5\% mIoU outperforming BYOL by 7.2\% mIoU and PixPro by 2.5\% mIoU in finetuning setting. In frozen setting, our method outperform BYOL by 6.4\% achieving 37.4\% mIoU.

% \sudipta{Please split up the text into two paragraphs, one for Cityscapes and one for ScanNet.}
% CityScapes \cite{cordts2016cityscapes} and ScanNet \cite{dai2017scannet} are very different from the PASCAL VOC and COCO datasets. On the other hand, CityScapes contain images from urban street scenes. Table \ref{tab:city_scane} reports semantic segmentation performance for both the CityScapes and ScanNet datasets. For both datasets, we use a FCN ResNet50-C4 network and finetune the network for 40k iterations. On CityScapes, our method achieves 6.2\% mIoU improvement over BYOL. On ScanNet, we outperform BYOL by 7.2\% mIoU. We achieve state-of-the-art performance for both datasets.

\begin{table}[!htbp]
    % \vspace{-4mm}
    \centering
        \centering
        \begin{tabular}{l  c | c | c }
            \toprule
            \multirow{2}{*}{\bf Method} & {\bf Pretrain} & {\bf VOC} & {\bf CityScapes}
            \\
                                        & {Epochs}       & {mIoU} & {mIoU}    \\
            \midrule
            % Scratch             & 53.9       \\
            Scratch                     & -              & 40.7  & 63.5    \\
            %  DenseCL CC & 56.7 & 81.7 & 63.0 & 67.5 \\
            %  \midrule
            Supervised                  & 90             & 67.7     & 74.6 \\
            Moco v2                     & 200            & 67.5     & 74.5 \\ %68.9       \\
            BYOL                        & 300            & 63.3     & 71.6 \\ % 64.4       \\
            DenseCL                     & 200            & 69.4    & 69.4  \\
            PixPro$^{\dagger}$          & 400            & 71.1     & 77.2 \\
            \rowcolor{_teal3} \ours     & 400            & \bf72.1  & \bf77.8 \\
            \bottomrule
        \end{tabular}
    \vspace{-3mm}
    \caption{\small {\bf Evaluation on Semantic Segmentation using FCN ResNet-50 network on PASCAL VOC and CityScapes dataset.} ($^{\dagger}$): We use pretrained checkpoint released by the authors and finetune the full networks on the VOC dataset. All other scores are obtained from the respective papers.}
    \label{tab:main_segm}
    \vspace{-4mm}
\end{table}

\subsection{Analysis}
\label{sec:analysis}
\vspace{-2mm}
\customparagraph{Frozen Backbone Analysis.}
We also report detection and segmentation results for frozen backbone following \cite{goyal2019scalinggoyal,henaff2021efficientdetcon,xiao2021regionresim}. Training a linear classifier on a frozen backbone is a standard approach to evaluate self-supervised representation quality for image classification \cite{caron2020unsupervisedswav,chen2020bigsimclrv2,grill2020bootstrapbyol,he2020momentummoco}. We adopt the standard strategy in `frozen backbone' setting where we freeze the pre-trained ResNet50 backbone and only fine tune the remaining layers. Frozen backbone might be an ideal evaluation strategy because fine tuning the full network evaluates quality of representations along with initialization and optimization, whereas frozen backbone evaluates mostly the representation quality of the backbone \cite{goyal2019scalinggoyal,xiao2021regionresim}.
% \sudipta{Can we include DenseCL here and in the table? After PixPro, it is the next paper that is most related to our work.}

\begin{table*}[!htbp]
    % \vspace{-3mm}
    \centering
    \begin{adjustbox}{max width=\linewidth}
        \begin{tabular}{p{0.16\linewidth} | c c c | c c c| c | c}
            \toprule
            \multirow{2}{*}{\bf \small Method} & \multicolumn{3}{c|}{{\bf \small PASCAL VOC OD}} & \multicolumn{3}{c|}{\bf \small COCO OD} & {\bf \small VOC SS}   & {\bf \small Cityscapes SS}                                                                  \\
                                               & { $\text{AP}^b$}                                & { $\text{AP}^b_{50}$}                   & { $\text{AP}^b_{75}$} & { $\text{AP}^b$}           & { $\text{AP}^b_{50}$} & {$\text{AP}^b_{75}$} & {mIoU} & {mIoU} \\
            \midrule
            Supervised                         & 50.7                                            & 80.4                                    & 55.1
                                               & 30.3                                            & 50.0                                    & 31.3
                                               & 56.6                                            & 55.7                                                                                                                                                          \\
            BYOL                               & 52.4                                            & 81.1                                    & 57.5
                                               & 30.2                                            & 49.1                                    & 31.5
                                               & 55.7                                            & 55.6                                                                                                                                                          \\
            DenseCL                            & 50.9                                            & 79.9                                    & 55.0                  & 25.5                       & 43.6                  & 25.8                 & 63.0   & 58.5   \\
            PixPro                             & 53.5                                            & 80.4                                    & 59.7
                                               & 27.7                                            & 44.6                                    & 29.1
                                               & 60.3                                            & 58.2                                                                                                                                                          \\
            \ours                              & \bf55.1                                         & \bf82.6                                 & \bf61.7
                                               & \bf30.5                                         & \bf49.8                                 & \bf31.7
                                               & \bf63.4                                         & \bf60.7                                                                                                                                                       \\

            \bottomrule
        \end{tabular}
    \end{adjustbox}
    \vspace{-2mm}
    \caption{\small \textbf{ Frozen backbone evaluation.} {\small We freeze the ResNet-50 backbone and finetune the other layers (RPN, FPN, classifier networks, regression layers, etc.). We use faster-RCNN with FPN for PASCAL VOC object detection (OD), RetinaNet-RCNN for COCO detection (OD), and FCN network for VOC and CityScapes segmentation (SS). For this experiment, we use publicly available checkpoint for the backbone networks and evaluate  on the downstream tasks.}}
    \label{tab:frozen}
    \vspace{-2mm}
\end{table*}

For frozen backbone (Table \ref{tab:frozen}), we achieve 30.5\% AP outperforming PixPro by 2.8\% AP on COCO object detection, 55.1\% AP for VOC detection outperforming PixPro by 1.6\% and BYOL by 2.7\%. We achieve 63.4\% mIoU on VOC semantic segmentation, which is \textit{more than the score achieved by BYOL in finetuning setting (63.3\% mIoU)}. We also outperform PixPro by a significant 2.9\% mIoU. On CityScapes semantic segmentation, We achieve 60.7\% mIoU which improves upon BYOL by 5.1\% and PixPro by 2.5\%.

\customparagraph{Efficient Pre-training.}
% \todo{Show results for smaller image-size and epoch, and show that our method can outperform even 100 epoch BYOL model}
In Table \ref{tab:efficient}, we report results of VOC object detection with FasterRCNN-FPN, COCO object detection from MaskRCNN-FPN, VOC and CityScapes segmentation from FCN for BYOL and \ours pre-trained with different epochs. Results reveal that \textit{our model pre-trained with 200 epochs and with training image size 160 can achieve better results than BYOL pre-trained with 1000 epochs saving 5.3$\times$ computational resource}. Even our 100-epoch pre-trained model seems to be comparable with 1000-epoch pre-trained BYOL model. \textit{This validates efficacy of our local loss during self-supervised pre-training}.

% \vspace{-3mm}
\begin{table*}[!htbp]
    \vspace{-3mm}
    \centering
    \begin{adjustbox}{max width=\linewidth}
        \begin{tabular}{l c c | c | c | c | c | c}
            \toprule
            \multirow{2}{*}{\bf \small Method} & {\bf \small Pretrain} & {\bf \small Pretrain} & {\bf \small Pretrain} & \multicolumn{1}{c|}{{\bf \small VOC}} & \multicolumn{1}{c|}{\bf \small COCO} & {\bf \small VOC} & {\bf \small Cityscapes} \\
                                               & {\bf \small Epochs}   & {\bf \small Im-size}  & {\bf \small time}     & { $\text{AP}^b$}                      & { $\text{AP}^b$}                     & {mIoU}           & {mIoU}                  \\
            \midrule

            BYOL                               & 300                   & 224                   & $\times$1.6           & 56.9                                  & 40.4                                 & 63.3             & 71.6                    \\
            BYOL                               & 1000                  & 224                   & $\times$5.3           & 57.0                                  & 40.9                                 & 69.0             & 73.4                    \\
            \midrule
            \ours                              & 100                   & 224                   & $\times$0.8           & 58.2                                  & 40.9                                 & 68.4             & 76.5                    \\
            \ours                              & 200                   & 160                   & $\times$1             & 59.0                                  & 41.6                                 & 68.5             & 77.0                    \\
            \ours                              & 200                   & 224                   & $\times$1.6           & 59.6                                  & 42.0                                 & 70.9             & 77.4                    \\
            \bottomrule
        \end{tabular}
    \end{adjustbox}
    \vspace{-2mm}
    \caption{\small \textbf{Efficient SSL training on ImageNet.} Performance of object detection and segmentation for BYOL and \ours for different pre-training epochs and training image size. We achieve better performance than BYOL (1000 epochs pretraining) with our model pre-trained with 200 epochs and with training image size 160 that is 5.3$\times$ faster to pre-train.}
    \label{tab:efficient}
    \vspace{-1mm}
\end{table*}

\customparagraph{Importance of Local Contrast.}
% \ben{I think this paragraph should be called something like: ``The importance of local contrast''. We have a great opportunity to point out that global BYOL works better than global contrastive loss, but that local BYOL is much worse than Local Contrastive loss. Why is that? We need to see the standard arguments for BYOL vs. Contrastive and understand why they don't apply \emph{locally}. I think this would be more interesting than the ``DencseCL-like'' or ``max'' variants.}
In Table \ref{tab:contrast_over_mse}, we show relative performance of our local contrastive loss against non-contrastive BYOL-type loss. In, `BYOL+Local MSE loss', we apply the same L2-normalized MSE local loss as the global loss in BYOL. Models are trained for 200 epochs on the ImageNet dataset. We report the average AP scores for VOC detection, COCO detection with Mask-RCNN, and CityScapes segmentation, which shows that our approach of calculating local  consistency using contrastive loss works better than non-contrastive BYOL-type local loss.
% \begin{table}[!htbp]
%     % \vspace{-4mm}
%     \centering
%     \begin{minipage}[c]{0.4\textwidth}
%         \begin{adjustbox}{max width=\linewidth}
%             \begin{tabular}{c|c}
%                 \toprule
%                 {\small \bf Method} & {\small \bf VOC (AP)} \\
%                 \midrule
%                 BYOL                & 54.81                 \\
%                 % \ours (Random) & - \\
%                 \ours (BYOL-like)   & 55.98                 \\
%                 % \ours (DenseCL-like) & 55.85                 \\
%                 % \ours (max)          & 54.91                 \\
%                 \ours               & {\bf 57.22}           \\
%                 \bottomrule
%             \end{tabular}
%         \end{adjustbox}
%     \end{minipage}
%     \hspace{0.01\textwidth}
%     \begin{minipage}[c]{0.5\textwidth}
%         \caption{{ The effect of different variations of local loss function. Results are reported in terms of average AP for PASCAL VOC object detection with the faster-RCNN framework.}}
%         \label{tab:other_corr}
%     \end{minipage}
%     \vspace{-1mm}
% \end{table}

\begin{table}[!htbp]
    \vspace{-3mm}
    \centering
    \begin{adjustbox}{max width=0.9\linewidth}
        \begin{tabular}{l | c | c | c | c}
            \toprule
            \multirow{2}{*}{\bf \small Method} &   \multicolumn{1}{c|}{{\bf \small VOC}} & \multicolumn{1}{c|}{\bf \small COCO} & {\bf \small VOC} & {\bf \small Cityscapes} \\
                                               &  { $\text{AP}^b$}                      & { $\text{mAP}$}                     & {mIoU}           & {mIoU}                  \\
            \midrule
            BYOL    &            57.0                                  & 40.9                                 & 69.0             & 73.4                    \\
            BYOL +Local MSE loss    &             58.7                                & {42.0}                                 & {70.5}             & {76.7}                    \\
            \ours      &             \bf59.6                                  & \bf42.5                                 & \bf72.1             & \bf77.8                    \\
            \bottomrule
        \end{tabular}
    \end{adjustbox}
    \vspace{-3mm}
    \caption{\small \bf Effectiveness of our LC-loss over MSE-loss.}
    \label{tab:contrast_over_mse}
    % \vspace{-1mm}
\end{table}

\customparagraph{Few-shot Image Classification.}
Since global and local losses appear to be complementary to each other, we ascertain if our method hurts the image classification performance for transfer learning. We use our pre-trained models as fixed feature extractors, and perform 5-way 5-shot few-shot learning on 7 datasets from diverse domains using a logistic regression classifier. Table \ref{tab:fshot} reports the 5-shot top-1 accuracy for the 7 diverse datasets. Table \ref{tab:fshot} reveals that \ours shows the best performance on average among the self-supervised models that use local consistency. \ours outperforms PixPro by 2.4\% top-1 accuracy on average; the minor fluctuation can be attributed to random noise.

\begin{table*}[!htbp]
    \vspace{-4mm}
    \centering
    \begin{adjustbox}{max width=\linewidth}
        \begin{tabular}{c|c c c c c c c | c}
            \toprule
            {\bf Method} & {\bf EuroSAT}\cite{helber2019eurosat} & {\bf CropDisease}\cite{mohanty2016cropdisease} & {\bf ChestX}\cite{wang2017chestx} & {\bf ISIC}\cite{codella2019skinisic} & {\bf Sketch}\cite{wang2019learningsketch} & {\bf DTD}\cite{cimpoi2014describingdtd} & {\bf Omniglot}\cite{lake2015humanomniglot} & {\bf Avg} \\
            \midrule
            Supervised   & 85.8                                  & 92.5                                           & 25.2                              & 43.4                                 & 86.3                                      & 81.9                                    & 93.0                                       & 72.6      \\
            \midrule
            SoCo         & 78.3                                  & 84.1                                           & 25.1                              & 41.2                                 & 81.5                                      & 73.9                                    & 92.2                                       & 68.0      \\
            DenseCL      & 77.7                                  & 81.0                                           & 23.8                              & 36.8                                 & 76.5                                      & 78.3                                    & 77.4                                       & 64.5      \\
            PixPro       & 80.5                                  & 86.4                                           & 26.5                              & 41.2                                 & 81.5                                      & 73.9                                    & 92.2                                       & 68.9      \\
            \ours        & 84.5                                  & 90.1                                           & 25.2                              & 41.9                                 & 85.6                                      & 80.2                                    & 91.5                                       & 71.3      \\
            \bottomrule
        \end{tabular}
    \end{adjustbox}
    \vspace{-2mm}
    \caption{\small {{\bf Few-shot learning results on downstream datasets.} The pre-trained models are used as fixed feature extractors We report top-1 accuracy for 5-way 5-shot averaged over 600 episodes. We use the publicly available pre-trained backbone as feature extractor for the few-shot evaluation.}}
    % \vspace{-2mm}
    \label{tab:fshot}
    % \vspace{-4mm}
\end{table*}

% \begin{figure}[!htbp]
%     \centering
%     \includegraphics[width=0.45\textwidth]{example-image-a}
%     \caption{Experiments on COCO-pretrained models.}
%     \label{fig:my_label}
% \end{figure}

% \customparagraph{Computational Complexity.}
% \ben{This is pretty weak as is and, without more information and comparison, should probably be dropped. Let's get a proper quantitative measure for ``complexity'' and talk about that instead.}Our \putalg has minimum overhead over the BYOL for the training time. To train our network for 200 epochs, it takes around 31 hour 30 min, whereas without the \putalg it takes 31 hour 17 min. Hence, it is just 1\% slower than BYOL. 

\customparagraph{Transfer to Other Vision Tasks.} Even though we mainly evaluted on detection and segmentation, we also show results for \textbf{keypoint estimation} a task that might benefit from models trained with local consistency. We use Mask-RCNN (keypoint version) with ResNet50 FPN network to evaluate keypoint estimation. We fine tune on COCO \texttt{train 2017} for 90k iterations. Table \ref{tab:coco_kp} shows that our method outperforms all other approaches in keypoint estimation task.

\begin{table}[!htbp]
    \vspace{-3mm}
    \centering
    % \begin{minipage}[c]{0.6\textwidth}
    \begin{adjustbox}{max width=0.9\linewidth}
        \begin{tabular}{l c c c c}
            \toprule
            {\bf Method}                               & {\bf Pretrain Epoch} & {\bf $\text{AP}$} & {\bf $\text{AP}_{50}$} & {\bf $\text{AP}_{75}$} \\
            \midrule
            % Scratch                 & -                    & -                 & -                      & -                      \\
            Supervised                                 & 90                   & 65.7              & 87.2                   & 71.5                   \\
            BYOL                                       & 300                  & 66.3              & 87.4                   & 72.4                   \\
            % \midrule
            VADeR \cite{pinheiro2020unsupervisedVADeR} & 200                  & 66.1              & 87.3                   & 72.1                   \\
            SCRL  \cite{roh2021spatiallyscrl}          & 1000                 & 66.5              & 87.8                   & 72.3                   \\
            DenseCL$^{\dagger}$                        & 200                  & 66.2              & 87.3                   & 71.9                   \\
            PixPro $^{\dagger}$                        & 400                  & 66.6              & \bf87.8                & 72.8                   \\
            \rowcolor{_teal3} \ours                    & 400                  & \bf67.2           & 87.4                   & \bf73.7                \\
            \bottomrule
        \end{tabular}
    \end{adjustbox}
    % \end{minipage}
    %
    % \hspace{0.01\linewidth}
    % \begin{minipage}[c]{0.35\textwidth}
    \vspace{-2mm}
    \caption{\small \textbf{COCO keypoint estimation.}
        Supervised and BYOL results are from \cite{roh2021spatiallyscrl}.
        ($^{\dagger}$) denotes We use the publicly available ImageNet-pretrained checkpoints released by the authors and finetune on the COCO dataset.}
    \label{tab:coco_kp}
    % \end{minipage}
    \vspace{-2mm}
\end{table}

\customparagraph{Detection on Mini COCO.}
As the full COCO dataset contains extensive annotated images for supervision, it might not always reveal the generalization ability of the network \cite{he2019rethinking}. We also report results for object detection on smaller versions of COCO training set in Table \ref{tab:coco_semi}. We report results when only 5\% and 10\% of the images (randomly sampled) are used for fine tuning the mask-RCNN with FPN network with 1$\times$ schedule. The evaluation is performed on the full \texttt{val2017} set. For the 5\% setting, our method outperforms BYOL by 1.4\% AP for the ImageNet pretrained models. For the 10\% setting, our method achieves improvement over BYOL by 1.8\% AP.

\begin{table}[!htbp]
    % \vspace{-3mm}
    \centering
        \begin{adjustbox}{max width=\linewidth}
            \begin{tabular}{p{0.26\linewidth} c  c  c  | c c c}
                \toprule
                                        & \multicolumn{3}{c}{\bf 5\%} & \multicolumn{3}{c}{\bf 10\%}                                                                                                \\
                \cmidrule(lr){2-4} \cmidrule(lr){5-7}
                {\bf Method}            & {\bf $\text{AP}$}           & {\bf $\text{AP}_{50}$}       & {\bf $\text{AP}_{75}$} & {\bf $\text{AP}$} & {\bf $\text{AP}_{50}$} & {\bf $\text{AP}_{75}$} \\
                \midrule
                % Scratch   \\
                Supervised              & 19.2                        & 31.0                         & 20.5                   & 25.0              & 39.9                   & 26.6                   \\
                BYOL                    & 21.9                        & 36.2                         & 23.2                   & 27.1              & 43.4                   & 29.3                   \\
                PixPro                  & 20.3                        & 31.4                         & 22.1                   & 25.4              & 39.5                   & 27.4                   \\
                \rowcolor{_teal3} \ours & \bf23.3                     & \bf37.4                      & \bf25.0                & \bf27.9           & \bf44.0                & \bf30.1                \\
                \bottomrule
            \end{tabular}
                        \end{adjustbox}

        \vspace{-2mm}
        \caption{\small {{\bf Object detection on mini-COCO with 1$\times$ schedule.} All scores are obtained from finetuning the publicly available pretrained backbone on the downstream dataset.}}
        \vspace{-2mm}
        \label{tab:coco_semi}

    \vspace{-3mm}
\end{table}

\customparagraph{Generalization to other SSL methods. } In the Appendix, we show results of our approach applied on other SSL approaches (e.g., DINO), where we also show consistent improvement over baseline methods.

\subsection{Ablation Studies} \label{sec:ablation}
\begin{figure}[!htbp]
    \vspace{-2mm}
    \centering
    \begin{subfigure}{0.45\textwidth}
        \centering
        \includegraphics[width=0.48\linewidth]{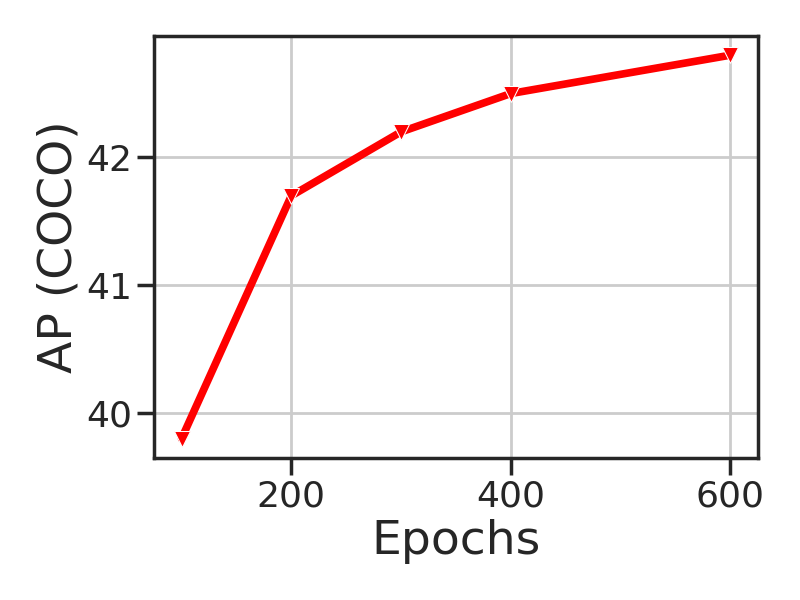}
        \includegraphics[width=0.48\linewidth]{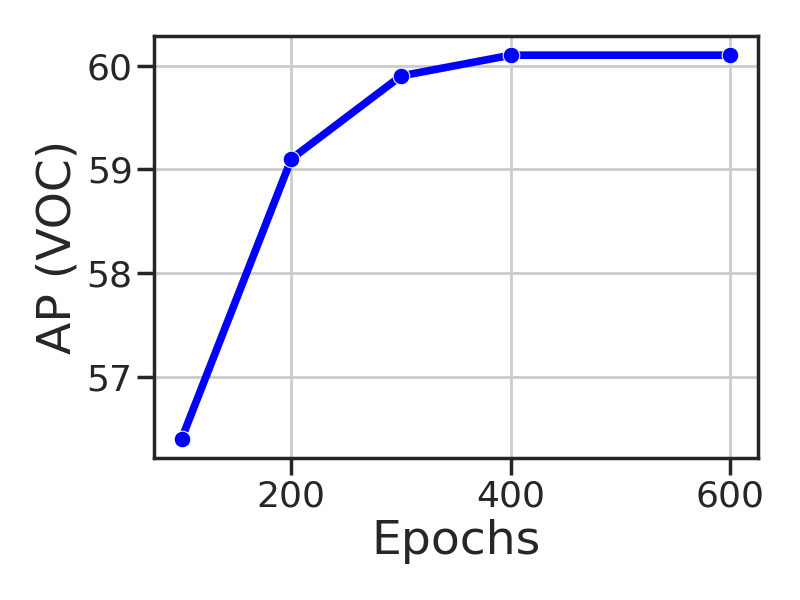}
        \vspace{-3mm}
        \caption{\small {\bf  Effect of longer pre-training.} The plot shows average AP for object detection on COCO validation sets (left) and PASCAL VOC (right).}
        \label{fig:abl_ep}
    \end{subfigure}
    \hfill
    \begin{subfigure}{0.4\textwidth}
        \centering
        \includegraphics[width=0.6\textwidth]{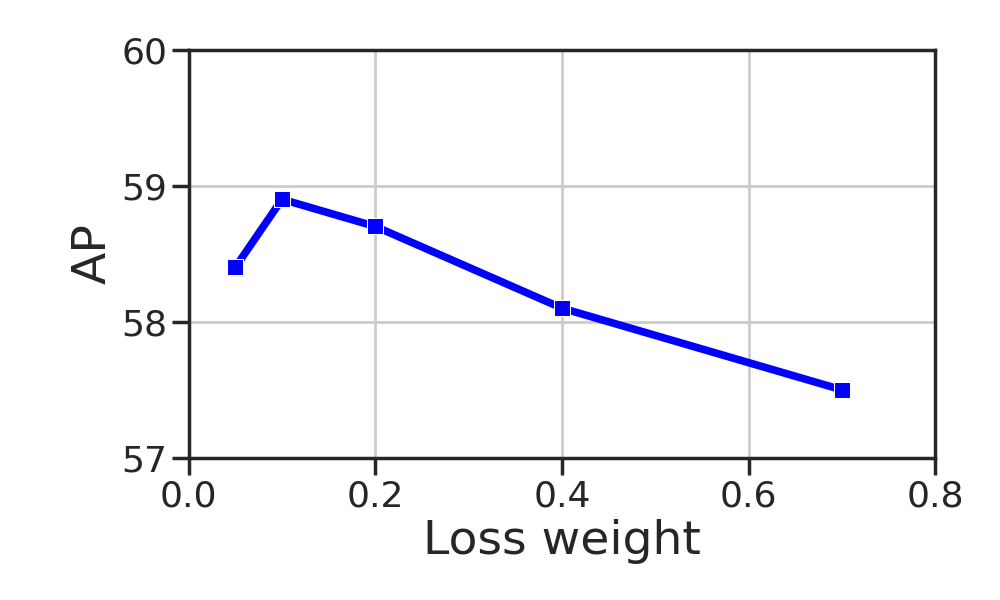}
        \caption{\small {\bf  Effect of relative weights of \putalg}. The plot shows the average AP for object detection on VOC.}
        \label{fig:abl_alpha}
    \end{subfigure}
    \vspace{-2mm}
    \caption{\small {\bf Ablation Studies} on pretraining epochs and relative weights of local contrastive loss.}
    \vspace{-3mm}
\end{figure}

\vspace{-1mm}
\customparagraph{Effect of Pretraining Epochs.}
Figure \ref{fig:abl_ep} reports object detection performance on PASCAL VOC with faster-RCNN-FPN and MS-COCO with Mask-RCNN-FPN for different numbers of pre-training epochs. The models are pre-trained on the ImageNet training set. Longer training generally results in better downstream object detection performance. For example, for the 100 epoch pre-trained model, the AP is 39.8\%, whereas for the 600 epoch pre-trained model it improves to 42.8\% for COCO evaluation. Upswing is also observed for PASCAL VOC object detection.

% \begin{figure}[!htbp]
%     % \vspace{-2mm}
%     \centering
%     \begin{minipage}[c]{0.6\textwidth}
%     \includegraphics[width=0.45\linewidth]{Figures/abl_ep_coco.png}
%     \includegraphics[width=0.45\linewidth]{Figures/abl_ep_pascal.png}    
%     \end{minipage}
%     \begin{minipage}[c]{0.3\textwidth}
%     \caption{{\bf \small Effect of longer training.} The plot shows average AP for object detection on PASCAL VOC (left) and COCO validation sets (right). The models were pretrained on the COCO dataset.}
%     % \vspace{-3mm}
%     \label{fig:abl_ep}
%     \end{minipage}
% \end{figure}

\customparagraph{Ablation on Loss Weight {\bf $\alpha$}.}
% \ben{Can we offer better anaylsis than this? Clearly the numbers suggest that a balance between global and local losses gets the best local performance. Can we do anything more in depth here?}
Figure \ref{fig:abl_alpha} reports the AP for object detection on PASCAL VOC for different values of the weight parameter $\alpha$. The models are pre-trained on the ImageNet dataset for 200 epochs with training image size of 160 for faster training. $\alpha$ balances the weight between the global and local loss functions. For $\alpha=0.05$, the mean AP is 58.4\%. We get a slightly better performance with $\alpha=0.1$ (58.9\%) and $\alpha=0.3$ (59.0\%). The performance degrades a little when $\alpha$ is increased to $\alpha=0.7$ (57.5\%). Results reveal the best performance is achieved when we use both global and local loss functions, and a proper balance between them ensures better downstream performance.

\subsection{Qualitative Analysis}
\begin{figure}[!tbp]
    \centering
    \includegraphics[width=0.3\linewidth]{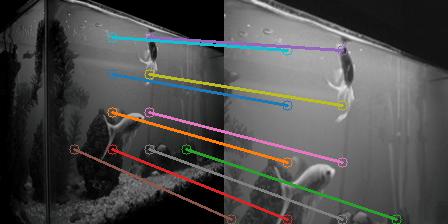}
    \includegraphics[width=0.3\linewidth]{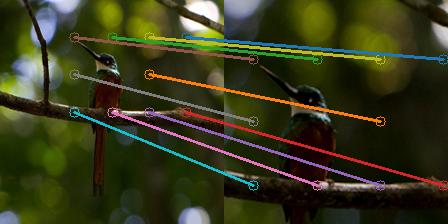}
    \includegraphics[width=0.3\linewidth]{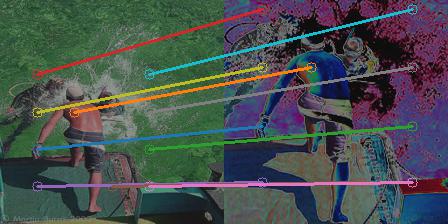}
    \\
    \includegraphics[width=0.3\linewidth]{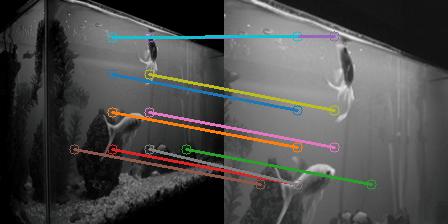}
    \includegraphics[width=0.3\linewidth]{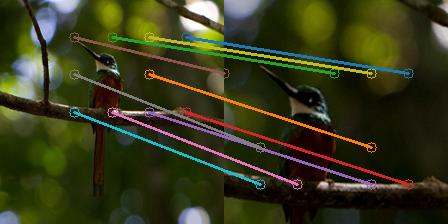}
    \includegraphics[width=0.3\linewidth]{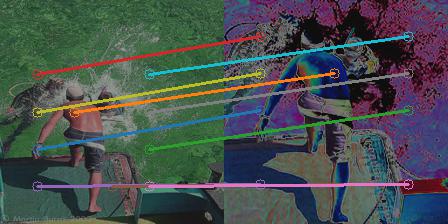}
    \vspace{-2mm}
    \caption{\small { {\bf Correspondence Visualization.} (Top) Ground-truth correspondence between two transformed versions of the same image. (Bottom) Correspondence prediction from \ours. }}
    \vspace{-2mm}
    \label{fig:vis_corr}
\end{figure}

\vspace{-1mm}
\customparagraph{Correspondence Visualization.}
In Figure \ref{fig:vis_corr}, we show visual examples of correspondence from our model. For two transformed images $I_1$ and $I_2$, we extract feature representations $F_1$ and $F_2$. For each feature in $F_1$, the corresponding feature in $F_2$ is calculated based on maximum cosine similarity between the feature representations. We show the matching at the original image resolution. Figure \ref{fig:vis_corr} shows that our method predicts accurate matches most of the time (considering the resolution error due to the $32\times 32$ grid size in the pixel space for each feature point).

\begin{figure}[!tbp]
    \centering
    % \begin{subfigure}[c]{0.32\linewidth}
    %     \centering
    %     \includegraphics[width=0.95\linewidth]{Figures/store_plt_sym/0_0000_o.jpg}
    %     \includegraphics[width=0.95\linewidth]{Figures/store_plt_sym/0_0008_o.jpg}
    %     \includegraphics[width=0.95\linewidth]{Figures/store_plt_sym/0_0026_o.jpg}
    %     \caption{GT}
    % \end{subfigure}
    \begin{subfigure}[c]{0.32\linewidth}
        \centering
        \includegraphics[width=0.95\linewidth]{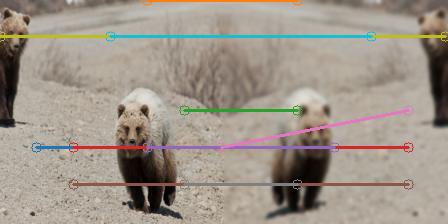}
        \includegraphics[width=0.95\linewidth]{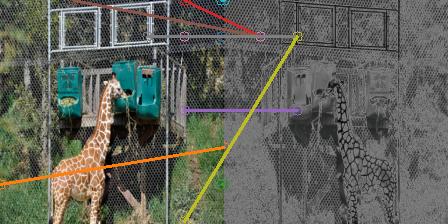}
        \includegraphics[width=0.95\linewidth]{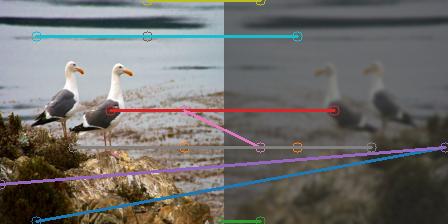}
        \caption{BYOL}
    \end{subfigure}
    \begin{subfigure}[c]{0.32\linewidth}
        \centering
        \includegraphics[width=0.95\linewidth]{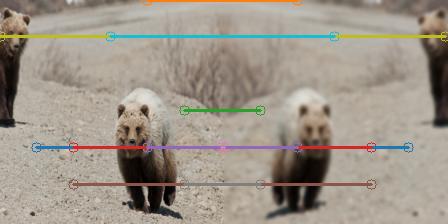}
        \includegraphics[width=0.95\linewidth]{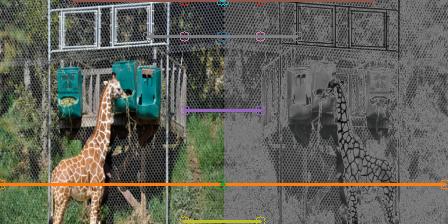}
        \includegraphics[width=0.95\linewidth]{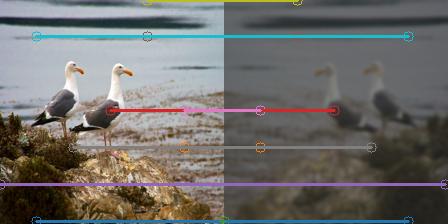}
        \caption{PixPro}
    \end{subfigure}
    \begin{subfigure}[c]{0.32\linewidth}
        \centering
        \includegraphics[width=0.95\linewidth]{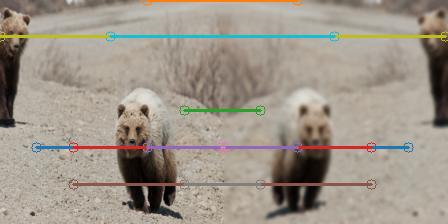}
        \includegraphics[width=0.95\linewidth]{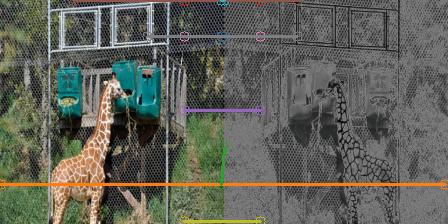}
        \includegraphics[width=0.95\linewidth]{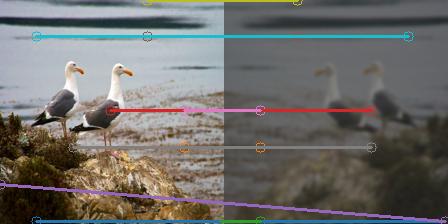}
        \caption{\ours}
    \end{subfigure}
    \vspace{-3mm}
    \caption{\small {\bf Correspondence map for an image and its transformed versions with only flipping and color transformation.} (a) Results from BYOL without local loss. There are many erroneous corresponding pixels denoting that global loss alone does not learn good features for local correspondence. (b) and (c): Results from PixPro and \ours, both of which perfectly detects most of the correspondences.}
    % \vspace{-3.5mm}
    \label{fig:more_corr}
    \vspace{-4mm}
\end{figure}

\customparagraph{More Analysis on Correspondence.}
To show that our method is learning better correspondence across datasets, we perform a simple experiment. Given an image, we flip the image along the horizontal direction, and apply color transformations (Gaussian blur, color jitter, and random grayscale operations). Since the only spatial transformation is horizontal flipping, the corresponding pixels are simply at the mirror locations of the original pixels, i.e., for a pixel location (x,y), the correct correspondence location in the transformed image is (w-x, y), where w is the width. Note that the correspondence is measured in the feature locations, not in actual pixel locations. We can also measure the correspondence accuracy based on whether the matching is correct or not. We use the ImageNet pre-trained backbone from BYOL and \ours, and evaluate the correspondence on the COCO dataset. Figure \ref{fig:more_corr} shows some visual examples of the correspondence map on images from the COCO dataset. We also measure the accuracy of correct correspondences on the COCO val dataset. For the BYOL pre-trained model, the accuracy is only $\sim$33\%, PixPro achieves $\sim$99\% accuracy, and \ours achieves $\sim$96\% accuracy. The results suggest that our approach is more robust against color transformations. We infer that PixPro achieves better accuracy as PixPro is trained only with local consistency loss, whereas we use both global and local correspondence during pre-training. The models have not been trained on COCO. Hence, the results also show that the correspondence maps generalize to other datasets.

\vspace{-1mm}
\section{Discussion and Conclusion}
\vspace{-2mm}
% \todo{Ashraful write this as a paragraph about globally consistent local representations.}

% Important point: let's not just make this a rehash of the introduction. Let's actually conclude something. Some conclusions that we can draw based on our experiments:
% \begin{itemize}
%     \item Can retain both spatial and global information in a single network.
%     \item Good global representation is important to having good local representations.
%     \item Smart augmentation schemes can provide us with local information ``for free''.
% \end{itemize}
% \ben{We need to rewrite this to reflect our updated angle.}

Even though our model consistently improves performance on detection and segmentation, it has some limitations. First, we calculate local correspondence loss at low spatial resolution (32$\times$ down-sampled from the original image resolution for ResNet50). Computing \putalg at higher resolutions may be beneficial but is much more computationally expensive. Thus, the trade-off between performance and accuracy needs to be further studied. Second, the local correspondences are not sampled at \textit{good} feature points (e.g., corners); rather they are sampled on a uniform 2D grid. Thus, \putalg might be too strict when dealing with large, texture-less image regions. Our loss also does not account for presence of self-similar image regions and the effect of not modeling them needs to be evaluated. 

To summarise, we propose a simple framework for self-supervised learning that leverages known pixel correspondences between different transformations of an image. We showed that the model pre-trained with our approach provides better representations for detection and segmentation. Imposing our loss enables a single network to retain both spatial and global information, both of which we have shown are necessary to obtain good features. Our training does not require any external supervision, since all the local and global constraints are generated from the input image itself. We showed that our method outperforms existing self-supervised methods that impose local consistency without requiring complex architectural components such as encoder-decoder layers, propagation modules, and regions proposal networks.

{\small
    \bibliographystyle{ieee_fullname}
    \bibliography{main}
}

\onecolumn
\appendix
{
    \section*{\LARGE{Appendix}}
}

The supplemental material contains additional analysis, and ablation studies.  All these are not included in the main paper due to the space limit.

\section{More Results}

\subsection{Results with Longer Finetuning}
Table \ref{tab:coco_2x} shows results with Mask-RCNN with FPN for 2$\times$ schedule on COCO dataset for ImageNet pretrained model. \ours improve over ReSim pretrained backbone by 1.6\% AP on COCO detection and 1.4\% AP on COCO segmentation.

\begin{table}[!htbp]
    \centering
    \begin{tabular}{l c c c c c c c}
        \toprule
        \multirow{2}{*}{\bf Method}                    &
        \multirow{2}{*}{\bf \begin{tabular}[c]{c}Pretrain\\ epoch\end{tabular}} &
        \multicolumn{3}{c}{\bf Detection}              &
        \multicolumn{3}{c}{\bf Instance-seg.}                                                                                                                \\
                                                       &     & AP$^{bb}$     & AP$^{bb}_{50}$ & AP$^{bb}_{75}$ & AP$^{mk}$ & AP$^{mk}_{50}$ & AP$^{mk}_{75}$ \\ \midrule
        Scratch                                        & -   & 36.7          & 56.7           & 40.0           & 33.7      & 53.8           & 35.9           \\
        Supervised                                     & 90  & 40.6          & 61.3           & 44.4           & 36.8      & 58.1           & 39.5           \\
        MoCo v2                                        & 200 & 40.9          & 61.5           & 44.6           & 37.0      & 58.4           & 39.6           \\
        ReSim-FPN$^{T}$                                & 400 & 41.9          & 62.4           & 45.9           & 37.9      & 59.4           & 40.6           \\
        \rowcolor{_teal3} \ours                        & 400 & \textbf{43.5} & \textbf{63.7}  & \textbf{47.5}  & \bf39.3   & \bf61.3        & \bf42.1        \\
        \bottomrule
    \end{tabular}
    \caption{\small \textbf{Object detection and instance segmentation on COCO for finetuning with 2$\times$ schedule.} We use Mask-RCNN with FPN for finetuning.}
    \label{tab:coco_2x}
\end{table}

% \subsection{Importance of Local Contrast}
% In Table \ref{tab:other_corr_supp}, we show relative performance of our local contrastive loss against non-contrastive BYOL-type loss. In, \ours (BYOL-like), we apply the same L2-normalized MSE local loss as the global loss in BYOL. Models are trained for 200 epochs on the ImageNet dataset. We report the average AP scores for PASCAL VOC in Table \ref{tab:other_corr}, which shows that our approach of calculating local  consistency using contrastive loss works better than non-contrastive BYOL-type local loss.
% \begin{table}[!htbp]
%     \centering
%     \begin{adjustbox}{max width=\linewidth}
%         \begin{tabular}{c|c}
%             \toprule
%             {\small \bf Method} & {\small \bf VOC (AP)} \\
%             \midrule
%             BYOL                & 55.0                  \\
%             % \ours (Random) & - \\
%             \ours (BYOL-like)   & 58.7                  \\
%             \ours               & {\bf 59.6}            \\
%             \bottomrule
%         \end{tabular}
%     \end{adjustbox}

%     \caption{{ The effect of different variations of local loss function. Results are reported in terms of average AP for PASCAL VOC object detection with the faster-RCNN with FPN framework. The models are pretrained on ImageNet dataset for 200 epochs.}}
%     \label{tab:other_corr_supp}
% \end{table}

\subsection{COCO Object and Instance Segmentation}

In Table \ref{tab:coco_mrcnn_supp}, we show more results of SSL methods on COCO object detection and instance segmentation using Mask-RCNN with FPN network for 90k iterations. Note that SoCo (4-views) \cite{wei2021aligningsoco} uses an additional views for pretraining to boost the performance.

\begin{table}[!htbp]
    \centering
    \begin{tabular}{l c c c c c c c}
        \toprule
        \multirow{2}{*}{\bf Method}                    &
        \multirow{2}{*}{\bf \begin{tabular}[c]{c}Pretrain\\ epoch\end{tabular}} &
        \multicolumn{3}{c}{\bf Detection}              &
        \multicolumn{3}{c}{\bf Instance-seg.}                                                                                                                       \\
                                                       &      & AP$^{bb}$      & AP$^{bb}_{50}$ & AP$^{bb}_{75}$ & AP$^{mk}$      & AP$^{mk}_{50}$ & AP$^{mk}_{75}$ \\ \midrule
        Scratch                                        & -    & 31.0           & 49.5           & 33.2           & 28.5           & 46.8           & 30.4           \\
        Supervised                                     & 90   & 38.9           & 59.6           & 42.7           & 35.4           & 56.5           & 38.1           \\
        \midrule
        MoCo                                           & 200  & 38.5           & 58.9           & 42.0           & 35.1           & 55.9           & 37.7           \\
        MoCo v2                                        & 200  & 40.4           & 60.2           & 44.2           & 36.4           & 57.2           & 38.9           \\
        InfoMin                                        & 200  & 40.6           & 60.6           & 44.6           & 36.7           & 57.7           & 39.4           \\
        BYOL                                           & 300  & 40.4           & 61.6           & 44.1           & 37.2           & 58.8           & 39.8           \\
        % SwAV                                           & 400  & -              & -              & -              & -              & -              & -              \\
        VADeR                                          & 200  & 39.2           & 59.7           & 42.7           & 35.6           & 56.7           & 38.2           \\
        \midrule
        ReSim-FPN$^{T}$                                & 200  & 39.8           & 60.2           & 43.5           & 36.0           & 57.1           & 38.6           \\
        % SoCo                                           & 400  & \bf43.0        & \boldsec{63.3} & \bf47.1        & 38.2           & \boldsec{60.2} & 41.0           \\
        SoCo                                           & 400  & 43.0        & {63.3} & 47.1        & 38.2           & {60.2} & {41.0}           \\
        SoCo (4-views)                                          & 400  & \bf43.2        & \bf{63.5} & \bf47.4        & \boldsec{38.4}           & \boldsec{60.2} & \boldsec{41.4}          \\
        DetCon$_{S}$                                   & 1000 & 41.8           & -              & -              & 37.4           & -              & -              \\
        DetCon$_{B}$                                   & 1000 & {42.7}         & -              & -              & 38.2           & -              & -              \\
        \midrule
        DenseCL                                        & 200  & 40.3           & 59.9           & 44.3           & 36.4           & 57.0           & 39.2           \\
        DetCo                                          & 800  & 40.1           & 61.0           & 43.9           & 36.4           & 58.0           & 38.9           \\
        InsLoc                                         & 400  & 42.0           & 62.3           & 45.8           & 37.6           & 59.0           & 40.5           \\
        PixPro                                         & 400  & 41.4           & 61.6           & 45.4           & -              & -              & -              \\
        \rowcolor{_teal3} \ours                        & 200  & 42.0           & 62.5           & 46.3           & 37.9           & 59.5           & 40.8           \\
        \rowcolor{_teal3} \ours                        & 400  & 42.5           & 62.9           & 46.7           & {38.3} & {60.0} & {41.1} \\
        \rowcolor{_teal3} \ours                        & 600  & \boldsec{42.8} & \boldsec{63.4}  & \boldsec{46.6} & \bf38.6        & \bf60.4        & \bf41.5        \\
        \bottomrule
    \end{tabular}
    \caption{\small \textbf{Object detection and instance segmentation fine-tuned on COCO.} We use Mask R-CNN R50-FPN (1$\times$ schedule), and report bounding-box AP (AP$^{bb}$) and mask AP (AP$^{mk}$).}
    \label{tab:coco_mrcnn_supp}
\end{table}

\subsection{VOC Object Detection}
In Table \ref{tab:main_pascal_supp}, we show additional results for PASCAL VOC object detection using faster-RCNN with FPN network. We use the pretrained models released by the authors and finetune on the downstream task using Detectron2 framework.

\begin{table*}[!htbp]
    \centering
    \begin{adjustbox}{max width=\linewidth}
        \begin{tabular}{l c | c c c}
            \toprule

            \multirow{2}{*}{\bf Method}                    &
            \multirow{2}{*}{\bf \begin{tabular}[c]{c}Pretrain\\ epoch\end{tabular}} &
            \multicolumn{3}{c}{\bf Detection}                                                                  \\

                                                           &     & AP$^{bb}$ & AP$^{bb}_{50}$ & AP$^{bb}_{75}$ \\
            \midrule
            Supervised  \cite{he2016deepresnet}            & 90  & 53.2      & 81.7           & 58.2           %& 55.4              & 82.1                   & 60.9                      \\
            \\
            Moco v2   \cite{chen2020improvedmocov2}        & 200 & 55.6      & 81.3           & 61.8           \\
            BYOL  \cite{grill2020bootstrapbyol}            & 300 & 55.0      & 83.1           & 61.1
            %& 56.9              & 83.4                   & 64.2               \\
            \\
            \midrule
            DetCo        \cite{xie2021detco}               & 800 & 56.7      & 82.3           & 63.0           \\
            ReSim-FPN \cite{xiao2021regionresim}           & 200 & 57.8      & 82.7           & 65.4           \\
            SCRL \cite{roh2021spatiallyscrl}               & 800 & 57.2      & 83.8           & 63.9           \\
            SoCo    \cite{wei2021aligningsoco}             & 400 & 57.4      & 82.6           & 64.7           \\
            \midrule
            DenseCL       \cite{wang2021densecl}           & 200 & 56.6      & 81.8           & 62.9           \\
            PixPro   \cite{xie2021propagatepixpro}         & 400 & 58.7      & 82.9           & 65.9           \\
            \rowcolor{_teal3} \ours                        & 400 & \bf60.1   & \bf84.2        & \bf67.8        \\
            \bottomrule
        \end{tabular}
    \end{adjustbox}
    \caption{\small \textbf{Object detection on PASCAL VOC.} We use faster RCNN with FPN for VOC object detection. We finetune all the layers including the pretrained backbone.}
    \label{tab:main_pascal_supp}
\end{table*}

\subsection{COCO Keypoint Esitmation.}
In Table \ref{tab:coco_kp_supp}, we show more results for COCO keypoint estimation. We use the pretrained models released by the authors and finetune on the downstream task using Detectron2 framework.

\begin{table}[!htbp]
    \centering
    \begin{adjustbox}{max width=0.9\linewidth}
        \begin{tabular}{l c c c c}
            \toprule
            {\bf Method}            & {\bf Pretrain Epoch} & {\bf $\text{AP}$} & {\bf $\text{AP}_{50}$} & {\bf $\text{AP}_{75}$} \\
            \midrule
            % Scratch                 & -                    & -                 & -                      & -                      \\
            Supervised              & 90                   & 65.7              & 87.2                   & 71.5                   \\
            Moco v2                 & 200                  & 65.9              & 86.9                   & 71.6                   \\
            BYOL                    & 300                  & 66.3              & 87.4                   & 72.4                   \\
            % \midrule
            VADeR$^{*}$             & 200                  & 66.1              & 87.3                   & 72.1                   \\
            SCRL$^{*}$              & 1000                 & 66.5              & 87.8                   & 72.3                   \\
            ReSim-FPN               & 200                  & 66.6              & 87.4                   & 72.8                   \\
            % SoCo                    & 400                  & -                 & -                      & -                      \\
            % \midrule
            % DetCo                   & 800                  & -                 & -                      & -                      \\
            DenseCL                 & 200                  & 66.2              & 87.3                   & 71.9                   \\
            PixPro                  & 400                  & 66.6              & \bf87.8                & 72.8                   \\
            \rowcolor{_teal3} \ours & 400                  & \bf67.2           & 87.4                   & \bf73.7                \\
            \bottomrule
        \end{tabular}
    \end{adjustbox}
    \caption{\small \textbf{COCO keypoint estimation.} We use the publicly available ImageNet-pretrained checkpoints released by the authors and finetune on the COCO dataset with Keypoint R50-FPN network for 90k iteration. ($^{*}$) denotes scores from the original papers.}
    \label{tab:coco_kp_supp}
\end{table}

\section{Analyzing Vision Transformer Backbone}

We perform experiments using a vision transformer backbone to evaluate whether our method benefits transformer pretraining. We use DINO \cite{caron2021emergingdino} self-supervised learning framework with Swin Transformer \cite{liu2021swin} backbone to evaluate our approach on vision transformer. We use Swin-T variant which has similar parameters as ResNet50, please refer to the original paper \cite{liu2021swin} for more details on the Swin Transformer backbone. We use AdamW optimizer with base learning rate of 0.0005 for batch-size 256 and weight-decay of 0.04. We pretrain the model for 300 epochs on 16 GPUs with 64 batch-size per GPU. For Swin with our approach, we set weight parameter $\alpha =0.5$. We use similar image transformations as the other experiments in the main paper for pretraining. The backbones are trained on ImageNet training set. We tried to also pretrain the backbone on COCO dataset, however, the network did not converge. We use mmdetection framework \cite{mmdetection} to train and evaluate on the downstream dataset with Swin transformer backbone. For faster-RCNN and mask-RCNN, we use ResNet50-FPN variant, train the models with AdamW optimizer for learning rate 0.0001 and weight-decay 0.05 on 8 GPUs with batch-size 2 per GPU. For segmentation, we use mmseg framework \cite{mmseg2020}. We use UperNet \cite{xiao2018unifiedupernet} for segmentation on VOC, CityScapes, and ScanNet. We use AdamW optimizer with learning rate 0.00006 and weight-decay 0.01. We train on PASCAL VOC for 20k iterations, and 40k iterations on the other datasets.

In Table \ref{tab:swin}, we report results for the PASCAL VOC, MS COCO, CityScapes, and ScanNet datasets. For PASCAL VOC detection, we use a Faster-RCNN framework with ResNet-50 FPN backbone, and for segmentation, we use the UperNet \cite{xiao2018unifiedupernet} framework. \putalg improves detection performance by 1.6\% AP and segmentation results by 2.1\% mIoU.  For COCO object detection and instance segmentation, we use mask-RCNN with a ResNet-50 FPN backbone. Our method outperforms the baseline by 0.8\% in detection and 0.4\% in segmentation. We perform segmentation on the CityScapes and ScanNet datasets using the UperNet framework. Our approach increases the mIoU by 0.1\% for CityScapes and 1.0\% for ScanNet. More technical details about the finetuning setup are in the supplementary material. We note that the improvement for transformer is not as impressive as for the ResNet-50 backbone, which suggests that the self-supervised vision transformer might already have more spatial information in its feature representation than ResNet-50. This also aligns with the conclusion from \cite{caron2021emergingdino} that the vision transformer contains information related to scene layout in the features.

\begin{table*}[!htbp]
    \centering
    \begin{adjustbox}{max width=\linewidth}
        \begin{tabular}{l | c c c c| c c c c c c| c | c}
            \toprule
            {\bf \small Method}  & \multicolumn{4}{c|}{{\bf \small PASCAL VOC}} & \multicolumn{6}{c|}{COCO} & {\bf \small Cityscapes}  & {\bf \small ScanNet}                                                                                                                                                                       \\
                                 & {\bf $\text{AP}^b$}                          & {\bf $\text{AP}^b_{50}$}  & {\bf $\text{AP}^b_{75}$} & {\bf mIoU}           & {\bf $\text{AP}^b$} & {\bf $\text{AP}^b_{50}$} & {\bf $\text{AP}^b_{75}$} & {\bf $\text{AP}^m$} & {\bf $\text{AP}^m_{50}$} & {\bf $\text{AP}^m_{75}$} & mIoU & mIoU \\
            \midrule
            DINO(Swin)           & 51.4                                         & 80.4                      & 56.0                     & 73.6                 & 40.2                & 62.3                     & 43.9                     & 37.6                & 59.3                     & 40.4                     & 78.0 & 62.1 \\
            DINO(Swin) + \putalg & 53.0                                         & 81.0                      & 57.8                     & 75.7                 & 41.0                & 62.9                     & 44.8                     & 38.0                & 59.9                     & 40.9                     & 78.1 & 63.1 \\
            \bottomrule
        \end{tabular}
    \end{adjustbox}
    \caption{{{\bf Results for the Swin transformer backbone pretrained with the DINO framework.} The models are pretrained on the ImageNet1K dataset for 300 epochs. \putalg loss improves the baseline consistently across datasets and tasks.}}
    \label{tab:swin}
\end{table*}

\section{More Ablations}
In Table \ref{tab:abl_more}, we show more ablation studies on learning rate, and momentum for the target network update during pretraining. The evaluation is performed with similar settings for VOC object detection with faster-RCNN-FPN and on COCO object detection with mask-RCNN-FPN in terms of average AP. We use image-size 160 for Table \ref{tab:abl_mom}, and image-size 224 for Table \ref{tab:abl_lr}.

\begin{table}[!htbp]
    \centering
    \begin{subtable}[t]{0.3\textwidth}
        \centering\begin{tabular}{l| c c}
            \toprule
            {lr} & { VOC} & { COCO} \\
            \midrule
            0.3  & 58.9   & 42.0    \\
            0.5  & 59.6   & 42.0    \\
            1.0  & 59.2   & 41.7    \\
            \bottomrule
        \end{tabular}
        \caption{\small {\bf learning rate}.}
        \label{tab:abl_lr}
    \end{subtable}
    \begin{subtable}[t]{0.3\textwidth}
        \centering\begin{tabular}{l| c c}
            \toprule
            {Mom.} & { VOC} & { COCO} \\
            \midrule
            0.99   & 59.1   & 41.3    \\
            0.996  & 58.4   & 41.6    \\
            0.999  & 56.6   & 40.4    \\
            \bottomrule
        \end{tabular}
        \caption{\small {\bf Momentum}.}
        \label{tab:abl_mom}
    \end{subtable}
    \caption{\textbf{Ablation studies}}
    \label{tab:abl_more}
\end{table}

\section{Computational Complexity.}
Our \putalg has minimum overhead over the BYOL for the training time. To train our network for 200 epochs, it takes around 31 hour 30 min, whereas without the \putalg it takes 31 hour 17 min. Hence, it is just 1\% slower than BYOL. In terms of GFLOPS, our \putalg has a minimum overhead of 8.54 GFLOPS, whereas BYOL has a minimum overhead of 8.29 GFLOPS.

\section{More Results on Few-shot Image Classification.}
Table \ref{tab:fshot_supp} shows few-shot learning results for SUpervised, BYOL, PixPro and ours. We use our pre-trained models as fixed feature extractors, and perform 5-way 5-shot few-shot learning on 7 datasets from diverse domains using a logistic regression classifier. It reports the 5-shot top-1 accuracy for the 7 diverse datasets. Results show the the best ours achieves better image classification scores than PixPro. However, the best method for few-shot transfer learning learning is BYOL. Similar finding are also reported in \cite{islam2021broad}.

\begin{table*}[!htbp]
    \centering
    \begin{adjustbox}{max width=\linewidth}
        \begin{tabular}{c|c c c c c c c | c}
            \toprule
            {\bf Method} & {\bf EuroSAT}\cite{helber2019eurosat} & {\bf CropDisease}\cite{mohanty2016cropdisease} & {\bf ChestX}\cite{wang2017chestx} & {\bf ISIC}\cite{codella2019skinisic} & {\bf Sketch}\cite{wang2019learningsketch} & {\bf DTD}\cite{cimpoi2014describingdtd} & {\bf Omniglot}\cite{lake2015humanomniglot} & {\bf Avg} \\
            \midrule
            Supervised   & 85.8                                  & 92.5                                           & 25.2                              & 43.4                                 & 86.3                                      & 81.9                                    & 93.0                                       & 72.6      \\
            \midrule
            BYOL         & 88.3                                  & 93.7                                           & 26.5                              & 42.3                                 & 86.8                                      & 83.5                                    & 94.7                                       & 73.7      \\
            PixPro       & 80.5                                  & 86.4                                           & 26.5                              & 41.2                                 & 81.5                                      & 73.9                                    & 92.2                                       & 68.9      \\
            \ours        & 84.5                                  & 90.1                                           & 25.2                              & 41.9                                 & 85.6                                      & 80.2                                    & 91.5                                       & 71.3      \\
            \bottomrule
        \end{tabular}
    \end{adjustbox}
    \caption{{{\bf Few-shot learning results on downstream datasets.} The pre-trained models are used as fixed feature extractors We report top-1 accuracy for 5-way 5-shot averaged over 600 episodes. We use the publicly available pre-trained backbone as feature extractor for the few-shot evaluation.}}
    % \vspace{-2mm}
    \label{tab:fshot_supp}
\end{table*}

\end{document}